\pdfoutput=1
\documentclass[10pt]{article}
\usepackage[preprint]{decode-tmlr}
\usepackage{amsmath,amssymb,booktabs,graphicx}
\usepackage{array,longtable}
\usepackage{float}
\PassOptionsToPackage{hyperfootnotes=false}{hyperref}
\usepackage{hyperref,url}
\usepackage{xurl}
\usepackage[section]{placeins}
\AddToHook{cmd/subsection/before}{\FloatBarrier}
\AddToHook{cmd/subsubsection/before}{\FloatBarrier}
\hypersetup{hidelinks}
\makeatletter
\newcommand{\unmarkedfootnote}[1]{\begingroup
  \renewcommand\@makefntext[1]{\noindent##1}\footnotetext{#1}\endgroup}
\makeatother
\setcitestyle{aysep={},citesep={,},yysep={,}}
\usepackage[inline]{enumitem}
\setlist{nosep}

\title{DECODE: Decomposable Environments for\\COntrolled Sequential-Decision Evaluation}

\author{\name Diego Silva Prado \email diego@alphasynthesis.com \\
\addr Alpha Synthesis Labs
\AND
\name Bruno Caldas de Souza \email caldas@alphasynthesis.com \\
\addr Alpha Synthesis Labs
\AND
\name Tiago Amaral Penteado de Jesus \email tiago@alphasynthesis.com \\
\addr Alpha Synthesis Labs
\AND
\name Ricardo Guimar\~aes Herrmann \email herrmann@alphasynthesis.com \\
\addr Alpha Synthesis Labs
\AND
\name Cristiano Breuel \email cbreuel@alphasynthesis.com \\
\addr Alpha Synthesis Labs
\AND
\name Cristiano Bevitori Maffia De Oliveira \email cbmo@alphasynthesis.com \\
\addr Alpha Synthesis Labs}

\hypersetup{pdfauthor={Diego Silva Prado and coauthors},pdftitle={DECODE}}

\begin{document}
\maketitle
\unmarkedfootnote{The code is available at
\url{https://github.com/alpha-synthesis/decode-arxiv}\newline and archived, with configurations and evidence, at
\url{https://doi.org/10.5281/zenodo.24200270}.}

\begin{abstract} When a sequential decision system falls short of an optimum, its score
does not reveal which modifications would improve its decisions. DECODE is a benchmark
for testing such modifications through controlled experiments on planning and available
information. Its consumer-credit collections generator provides five partially observed
environments that share known dynamics matrices and vary a memory parameter and process
noise. The experiments change selected inputs or planning rules while keeping the rest
of the solver fixed. Each comparison specifies the solver's state inputs, response
inputs and planning horizon. On the deterministic base environment, which was selected
for a large gap between six-month and one-month planning with the hidden state,
extending planning from one to six months increases the scores of two gradient-boosted
solvers by 15 to 17 percentage points of the exact optimum, computed with the hidden
state and borrower traits. Even at this horizon, the best graded solver, whose repayment
response is fitted on logs, remains 15.6 points below that optimum. In a specified
intervention sequence, three changes to its state inputs and their evaluation during
search lower its score by 6.3 points, and replacing its fitted repayment model with the
true response, evaluated with hidden borrower traits, then raises it by 15.6, a net
recovery of nine of those points. Because the intermediate effects are conditional on
the repayment model and the order of the changes, they do not uniquely assign the gap to
separate causes. Within these limits, DECODE makes hypotheses about solver performance
testable through specified interventions under known dynamics and explicit information
conditions. \end{abstract}

\section{Introduction}
\label{sec:intro}

Benchmarks for sequential decision systems often summarize performance with a final score.
A diagnostic evaluation can then test how specific changes in planning or available
information affect policy value. Interpreting the resulting change in value requires stating what information
each reference receives and what decision rule it applies. DECODE provides
partially observed environments with known equations and specified reference
policies for these comparisons.
Each environment is a finite-horizon partially observable Markov decision
process (POMDP)~\citep{kaelbling1998pomdp} with time-invariant dynamics. In the
consumer-credit collections setting used here, a hidden account state evolves
under these dynamics. Monthly actions affect its evolution and repayment through
known response maps, while policy value is expected net cash up to the first
recovery within the evaluation horizon. Under partial observation, a solver's
belief is the probability distribution of the hidden state given everything it
has observed.
To evaluate changes within this setting, DECODE combines the generator, a common
policy-value evaluator and the reference policies into a measurement instrument.
Knowing the world's equations allows changes to a solver's information and
planning inputs to be evaluated against references computed through closed-form
or path-dependent calculations on deterministic members and Monte Carlo estimates
under process noise. Each reference's guarantees depend on its information inputs
and planning rule. In particular, the optimum with access to the hidden state
differs from the filter-based reference.

\begin{figure}[!ht]
\centering
\includegraphics[width=\linewidth]{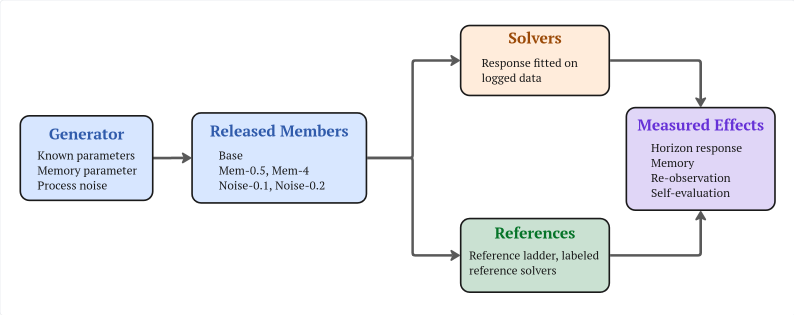}
\caption{Overview of DECODE.}
\label{fig:schematic}
\end{figure}

This paper proposes a reproducible instrument with specified input interventions and
reference policies, together with an empirical characterization of the released
environments. Its comparisons test controlled changes to information and planning inputs
for solvers fitted on partially observed collection logs. Graded solvers use logged
observations, while reference solvers receive specified input substitutions, including
ground-truth quantities. Both are evaluated under the same grading rule. The
environments vary along two dynamics parameters, a memory parameter and process noise,
which change the dependence on history and the uncertainty of future states. These
parameters define the five settings released as ``members'', with construction checks
(Fig.~\ref{fig:schematic}). The environments are decomposable in the sense that the
transition, the reading model and the repayment response are specified separately, so
each can be varied in the world or substituted into a solver one at a time. This
property does not imply an additive decomposition of a performance gap. This design
relates to diagnostic suites~\citep{osband2020bsuite,rajan2023mdpplayground}, generators
with certified optima on the full state~\citep{ibrahim2026converse} or on a sufficient
information state~\citep{ouerdane2026groundtruth}, and oracle-based analyses of
estimator errors~\citep{fu2019diagnosing}. The policy-value experiments are complemented
by predictive probes that test whether history improves prediction of the repayment
response. A separate comparison asks the corresponding predictive question for observed
outcomes in public loan-performance data. These predictive gains do not by themselves
establish improvements in policy value. The release supplies the generator, the five
members and the artifacts supporting these measurements.

On the member with the memory parameter and process noise both set to zero (Base),
extending the planning horizon from one month to six increases the score of the same
fitted solver by 15 to 17 points of the exact optimum. The construction checks of Base
require a gap of 15 to 40 points between the exact and one-month optima on a
training-world prefix and a best plan that varies across loans, so this setting is
selected to reward planning over the horizon and plans that differ across loans.
Contrasts between the tested solvers at a fixed horizon have point estimates below 4
points. Even at the longer horizon, the best graded solver, whose repayment response is
fitted on logs, remains 15.6 points below the exact optimum. In the declared
intervention sequence, three changes to the solver's state inputs and their evaluation
during search lower its score by 6.3 points, and supplying the world's response,
evaluated with the true hidden borrower traits, then raises it by 15.6, a net recovery
of nine of those points. The step effects are conditional on the fitted response and the
order of the changes. The step that replaces the history features with filter features
is small in that sequence, but the filter ignores repayment outcomes. Beyond these
changes in horizon and inputs, the benchmark checks how a planner evaluates and updates
its own plans. Chaining marginal repayment probabilities overstates the filter-based
planner's value relative to joint integration of the same plan under the same belief.
Allowing the planner to re-plan on fresh observations improves value on the noisy
members relative to executing its initial plan.

This paper is organized as follows. Section~\ref{sec:instrument} describes the
world generator, evaluation protocol, reference policies and solvers.
Section~\ref{sec:results} presents the experimental results.
Section~\ref{sec:discussion} discusses their interpretation and limitations.
Section~\ref{sec:related} reviews related work.
Section~\ref{sec:conclusion} concludes the paper.

\section{DECODE Benchmark Environment}
\label{sec:instrument}

This section explains how DECODE's environments, evaluation protocol and reference
calculations fit together in a diagnostic experiment. The world equations define
how the state evolves, what a solver observes and how actions affect repayment.
These equations govern the training logs and the trajectories used to evaluate
policies. The protocol specifies how the logs are collected and how policy value
is measured. Reference comparisons use this protocol to assess specified changes
in information and planning, while predictive probes measure how history
contributes to prediction.
Construction checks state the conditions for releasing the environments, and the
solver descriptions identify the models, information inputs and decision rules
compared under this protocol.

A world is a population of clients whose monthly outcomes depend on a hidden
state that evolves from one month to the next. In the consumer debt collection
setting used here, clients are delinquent loans, outcomes are repayments,
and actions are contact choices and settlement offers, each at a cost. A graded
solver never directly accesses the hidden state. It observes the public account
state and a noisy reading of the hidden state, and chooses an action per client
per month.

Two parameters vary these dynamics, the memory parameter $\delta$ and the process-noise
scale $\sigma$. Five settings are released as members. Base sets both parameters to
zero. Mem-0.5 and Mem-4 use $\delta = 0.5$ and $4$ with no process noise, while
Noise-0.1 and Noise-0.2 use $\sigma = 0.1$ and $0.2$ with $\delta = 0$. Mem-4 and
Noise-0.2 also use lower monthly base rates. The configurations are listed in
Table~\ref{tab:members} in Appendix~\ref{app:members}. Terms are defined at first use
and collected in the glossary of Appendix~\ref{app:glossary}.

\subsection{Dynamics and outcomes}
\label{sec:instrument-physics}

Each client carries a hidden state $z_t \in \mathbb{R}^6$ drawn from a standard Gaussian
population at origination and evolving once per month. An action $a_t = (c_t, d_t)$
pairs a contact channel $c_t$, one of none, short message service (SMS), WhatsApp, or
phone, with a settlement offer $d_t$. The channel enters the state transition through
contact intensities $w(c) = 0$, $0.4$, $0.4$, and $1$ respectively. SMS and WhatsApp
therefore move the state equally, at equal cost, and differ only in their effect on
repayment. With these intensities, the state follows a linear-Gaussian state-space
model~\citep{kalman1960} with an action-dependent rotation, given by: \begin{equation}
z_{t+1} \;=\; A\, Q_{w(c_t)}\, z_t \;+\; b\, w(c_t) \;+\; \eta_t , \qquad \eta_t \sim
\mathcal{N}(0, \sigma^2 I), \label{eq:transition} \end{equation} where $t$ indexes
months, $z_t \in \mathbb{R}^6$ is a client's hidden state and $z_{t+1}$ its next-month
state. The contact channel $c_t$ has intensity $w(c_t)$ as specified above. The matrix
$A \in \mathbb{R}^{6\times6}$ is a contraction and $b \in \mathbb{R}^6$ a fixed input
direction. The rotation $Q_w = \exp(w\,\delta\,S)$ uses channel intensity $w$, memory
parameter $\delta$ and a fixed skew-symmetric matrix $S \in \mathbb{R}^{6\times6}$, with
$\exp$ denoting the matrix exponential. The process innovation $\eta_t$ follows a
Gaussian distribution, denoted $\mathcal{N}$, with zero mean and covariance $\sigma^2
I$. Here $\sigma$ is the process-noise standard deviation per component and $I$ the
$6\times6$ identity matrix. The memory parameter $\delta$ scales the rotation in this
transition. At $\delta = 0$, $Q_w$ is the identity and the action adds the input
$b\,w(c)$, so the contributions of past actions add up independently and their order
matters only through how recently each was taken. At $\delta > 0$ each contact also
rotates the current state, which carries the inputs of all earlier contacts, so each
contact's effect depends on the state and on every earlier contact. This interaction is
what $\delta$ controls.

The process-noise scale $\sigma$, the second parameter, controls the monthly
innovations added to this evolution. When $\sigma = 0$ the state is a deterministic function of
the client's initial draw and the action history. At $\sigma > 0$ the state becomes
random, but each innovation is shrunk by $A$ in every later month, so the
accumulated variance converges to a finite level, at most
$\sigma^2/(1-\|A\|_2^2)$ per component of the hidden state, where
$\|A\|_2$ is the spectral norm of the transition matrix $A$ in
Eq.~\ref{eq:transition}.
The solver observes this evolving state through a two-dimensional linear
reading each month. The corresponding measurement equation is written as:
\begin{equation}
o_t \;=\; C\, z_t + \varepsilon_t, \qquad
\varepsilon_t \sim \mathcal{N}(0,\, 0.6^2 I_2),
\label{eq:observation}
\end{equation}
where $o_t \in \mathbb{R}^2$ is the reading at month $t$, $z_t$ is the hidden
state of Eq.~\ref{eq:transition}, and $C \in \mathbb{R}^{2\times6}$ is the fixed
loading matrix. The measurement noise $\varepsilon_t$ is zero-mean Gaussian with
standard deviation $0.6$ in each component, and $I_2$ is the $2\times2$ identity
matrix. Besides $o_t$,
the solver observes the public account state $x_t$ (delinquency ageing,
balance, contact fatigue, demographics), and the logged records also
carry realized repayments.
Repayment depends on the public account state as well as the hidden state and
borrower traits. Resilience, payment friction and contactability correspond to the three borrower
factors of \mbox{JUBILEE}~\citep{prado2026jubilee}, and regional economic stress to its
regional shock. DECODE combines
these mechanisms with the latent-state dynamics and observation model defined above.

Conditional on these model inputs, repayment is a monthly Bernoulli draw,
$r_t \sim \mathrm{Bernoulli}(\mathrm{expit}(y_t))$, where $r_t$ equals one for a
repayment and zero otherwise, $y_t$ is the logit of the repayment probability, and $\mathrm{expit}$ is the
logistic sigmoid. The corresponding logit is additive in the grouped terms, as in
logistic regression~\citep{cox1958}, and is given by:
\begin{equation}
y_t \;=\; \beta_0 + \gamma_{c_t} + g(x_t) + h_{c_t}(d_t)
  - \phi_{c_t}(x_t)\, f_t
  + \kappa\, u_{c_t}^{\top} z_t + \theta_{c_t}^{\top}\tau_i + \xi_t ,
\label{eq:logit}
\end{equation}
where $y_t$ is the repayment logit for borrower $i$ at month $t$, $c_t$ is the
contact channel and $d_t$ the settlement discount as a fraction of balance.
The subscript $c$ indexes channels.
The public account state $x_t$ is described after Eq.~\ref{eq:observation}, and
$z_t$ is the hidden state in Eq.~\ref{eq:transition}. The term
$\beta_0$ is the logit of the member's base rate, $\gamma_c$ a
channel offset, $g$ collects the public-state effects, $h_c$ is a
saturating response to the discount $d_t$, and $f_t$ is the contact-fatigue
component of $x_t$, with a sensitivity $\phi_c(x_t)$ that depends on
the borrower's observed income. Both $h_c$ and $\phi_c$ are the same for
every contact channel and vanish on no-contact months.

The remaining terms describe latent effects and outcome noise. The term
$\kappa\, u_{c_t}^{\top} z_t$ is the hidden score of the acting channel
$c_t$, with coupling vectors $u_c$ and scale $\kappa$. The term $\tau_i$ collects borrower $i$'s resilience, payment friction and
contactability, together with regional economic stress, with fixed weights
$\theta_c$. The superscript $\top$ denotes transpose, so the products with
$u_c$ and $\theta_c$ are inner products. Finally,
$\xi_t \sim \mathcal{N}(0, 0.2^2)$ is Gaussian logit noise with zero mean and
standard deviation $0.2$, drawn per loan-month. All additive terms are on the
log-odds scale. 
These repayment events also update the public account state. In the generator,
a recovery is a payment event and can leave the loan open.
It reduces the balance and resets delinquency to zero, returning the loan to
performing status. The loan closes as cured only if the remaining balance falls
below the larger of 50 currency units and 5\% of the original principal.
Charge-off is a competing terminal event that removes the loan without a recovery.

Besides their effect on repayment, two hidden traits affect how the historical
records are generated.
Payment friction steers the logging policy toward phone contact and away from no
contact, and contactability steers it the other way. Because they influence both action assignment and
repayment, they confound treatment and outcome. Graded solvers do not observe the
traits or $\xi_t$ directly. The oracle used for grading has access to the hidden
state and the equations.

The fixed matrices and vectors in these equations are shared by every client
of a world. One random draw of them is called a ``geometry''. It comprises $A$
and $b$ in the transition (Eq.~\ref{eq:transition}), the loading $C$ of the
reading (Eq.~\ref{eq:observation}), the coupling vectors $u_c$ of the hidden
score (Eq.~\ref{eq:logit}), and, on members with $\delta > 0$, the rotation
generator $S$. The released members share one geometry. Robustness sweeps examine
four further draws of these same components.

\subsection{Protocol}
\label{sec:instrument-protocol}

\begin{figure}[!ht]
\centering
\includegraphics[width=\linewidth]{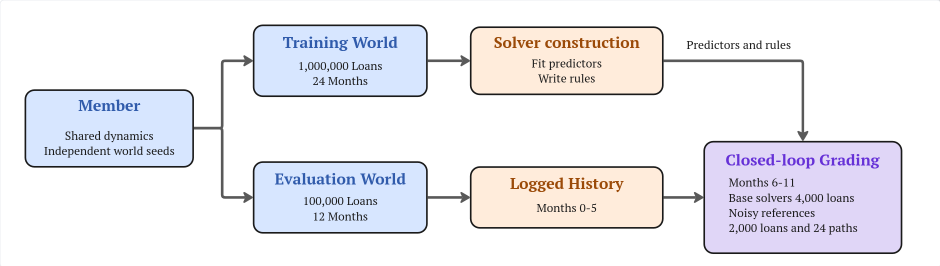}
\caption{World generation and policy-value evaluation.}
\label{fig:protocol}
\end{figure}

Training and evaluation worlds share each member's dynamics but use independent
seeds (Fig.~\ref{fig:protocol}). The training logs supply the data used to
construct graded solvers, while reference policies use the known world equations.
With probability $\varepsilon = 0.03$, logging
substitutes a uniformly drawn contact channel and discount. The no-contact action
is never randomized. Exploration excludes liquidation ages above 240 days, where
liquidation age counts days in the state of at least 90 days past due.
Decisions use a grid pairing four channels with 21 discount levels from zero to
half the balance, giving $(4 \times 21)^6$ six-month plans.

After the logged history, solvers take over at month six and re-decide monthly. The
harness supplies the earlier readings, channels, offers and contact fatigue, together
with the current account state. CE-Seq, the certainty-equivalent (CE) sequential solver
of Section~\ref{sec:instrument-solvers}, additionally loads earlier account states
directly from the evaluation database, revealing past repayments. Cohorts are selected
in loan-identifier order from loans open at the decision point, with the same fixed
cohort used for every graded solver. Counterfactual trajectories use common random
numbers across the compared policies. Under process noise, innovation paths are
independent across loans and shared across references.

Policy value is expected net cash up to the first recovery, with discounts
counted as lost and contact costs included. Monthly cash and costs are weighted
by the probability of no earlier recovery or charge-off. The grader evaluates
Eq.~\ref{eq:logit} at $\xi_t=0$ and integrates repayment probabilities without
drawing payments. During grading, solvers therefore receive readings and
deterministic account updates, without new realized payment outcomes. This
objective differs from the generator's complete cash flow, which can retain a
loan after payment.

\subsection{Reference ladder}
\label{sec:instrument-ladder}

A disclosure replaces one input of a graded solver with a specified reference
quantity. The resulting copy is labeled a reference solver and does not compete
with the graded solvers. Its change in value depends on how the fitted system
uses that input and can be negative. It differs from the decision-theoretic value
of free information for an optimal decision maker who can ignore
it~\citep{howard1966}. The policy-value references in Table~\ref{tab:references}
use the world's response equations and public account state under the declared
grading objective and action grid. The exact optimum is the fully observed
Markov decision process (MDP) upper bound~\citep{hauskrecht2000pomdp}.

\begin{table}[!ht]
\centering
\caption{Policy-value references.}
\label{tab:references}
\footnotesize
\setlength{\tabcolsep}{3pt}
\begin{tabular}{@{}>{\raggedright\arraybackslash}p{0.16\linewidth}>{\raggedright\arraybackslash}p{0.25\linewidth}>{\raggedright\arraybackslash}p{0.27\linewidth}>{\raggedright\arraybackslash}p{0.26\linewidth}@{}}
\toprule
Reference & Information & Calculation & Interpretation \\
\midrule
Exact optimum ($\sigma=0$) & True hidden state and borrower traits & Enumerate channel sequences and optimize discounts by backward dynamic programming & MDP upper bound for the grading objective \\
One-month optimum ($\sigma=0$) & True hidden state and borrower traits & Optimize the current month and re-decide monthly & Reference for myopic planning \\
Filter planner & Gaussian belief from readings, with true borrower traits & Integrate monthly repayment probabilities and chain their marginals & Value of a specified policy, without POMDP optimality \\
True-state re-planning ($\sigma>0$) & Current true hidden state and borrower traits & Monthly re-planning, evaluated over shared innovation paths & Normalizer, not an upper bound \\
\bottomrule
\end{tabular}
\end{table}

Without process noise, the state path is fixed by the initial state and channel
sequence. Discounts do not alter that path, allowing the search to enumerate $4^6$
channel sequences and optimize their discount grids separately. The reference filter
instead forms a Gaussian reading-model belief by a Kalman recursion~\citep{kalman1960},
using action-dependent transitions on the memory members. It treats logged actions as
fixed inputs and ignores repayment outcomes. Since the logging policy reads delinquency,
actions can carry outcome information, and this belief differs from the posterior given
the full public history. The filter planner masks future readings and neglects the
effect of actions on future information~\citep{barshalom1974dual}. Under process noise,
future innovation variance enters both the belief-based and true-state references. The
search integrates each month's probability over the predictive distribution and chains
its marginals, omitting cross-month dependence. Scores are percentages of each member's
upper reference, the exact optimum without process noise and true-state monthly
re-planning under noise. Under noise this reference normalizes scores and is not an
upper bound. At the path count in Fig.~\ref{fig:protocol}, each reference's estimated
Monte Carlo standard error is below 0.5\% of its normalizer. The standard errors are
themselves estimated and quantify Monte Carlo precision. They do not certify policy
optimality.

Predictive probes measure how past readings and actions inform the repayment response
(Table~\ref{tab:probes}). Each score is the out-of-fold $R^2$, the coefficient of
determination over three folds, of a gradient-boosted regressor~\citep{friedman2001gbm}
fitted to the true repayment logit $y_t$. Rows are phone-contact months among the
exploration rows with a complete six-month window, fixing the acting channel within
randomized assignment. Account fields enter at their start-of-month values, before the
repayment outcome.

\begin{table}[!ht]
\centering
\caption{Information probes and contrasts.}
\label{tab:probes}
\footnotesize
\begin{tabular}{@{}lp{0.60\linewidth}@{}}
\toprule
Probe & Information set \\
\midrule
Snapshot   & Current month's public state, reading, and offered discount \\
History    & Snapshot set plus the last six months of readings and actions (recorded by channel) \\
True state & Snapshot set plus the hidden state $z_t$ \\
\midrule
Contrast & What it measures \\
\midrule
Separation, history $-$ snapshot & Increase in fitted predictive $R^2$ from the six-month window \\
Headroom, true state $-$ history & Increase in fitted predictive $R^2$ from adding the true state \\
\bottomrule
\end{tabular}
\end{table}

A separate information reference uses the residual hidden-score variance for a predictor
that knows all other logit terms. It is written as: \begin{equation} R^2_{\mathrm{ref}}
\;=\; 1 \;-\; \frac{\operatorname{Var}\!\left(\kappa\, u_c^{\top}(z_t - \mu_t)\right)}
{\operatorname{Var}(y_t)}. \label{eq:bound} \end{equation} where $R^2_{\mathrm{ref}}$ is
the information reference and $\operatorname{Var}$ denotes variance across the probes'
measurement rows. The repayment logit $y_t$, hidden-score scale $\kappa$, coupling
vector $u_c$ and transpose notation are defined after Eq.~\ref{eq:logit}. Here $c$ is
the phone channel, $z_t$ is the hidden state at month $t$, and $\mu_t$ is its
reading-model posterior mean with logged actions treated as fixed inputs. The numerator
is the variance of the residual hidden score, and the denominator is the total logit
variance.

The probes use entering account states and logged actions that can reveal past
repayments omitted by this reading-model calculation. Conversely, the information
reference treats hidden traits and logit noise as known, although the probes cannot
observe them. It therefore does not bound or validate the probes, and crossing it is not
by itself evidence of an implementation fault.

\subsection{Checks}
\label{sec:instrument-gates}

The released environments are subject to construction checks on recovery rates
and information measures. Recovery-rate bands require the share of loans
recovered within one month and within nine months to fall inside pre-set ranges.
These checks are applied to the training and evaluation worlds of the four
members other than Base and to a 100{,}000-borrower prefix of Base's training
world. The three members with a deterministic reference additionally have
information clauses on the probes and values (Appendix~\ref{app:members}). These
are construction criteria, rather than external validation of collection behavior.

Check criteria were fixed before generation. One information clause was amended once
before release, and three members were regenerated with an adjusted base rate
(Appendix~\ref{app:members}). The release carries each member's check report and
regeneration manifest. The reports record passing checks for the released
configurations.

\subsection{Solvers}
\label{sec:instrument-solvers}

The graded solvers in Table~\ref{tab:solvers} combine logged information with a decision
rule. GBM-lags and GBM-flat use calibrated gradient-boosted machines
(GBMs)~\citep{friedman2001gbm} to predict repayment. Their charge-off hazard, cash
formula and fatigue dynamics are supplied by the generator, so only the repayment
response is fitted. The disclosures use GBM-lags. All other solvers except LLM-agent, a
rule set written by a large language model (LLM), also fit a predictor from the training
logs.

A myopic solver maximizes current-month expected value, whereas a sequential
solver uses the channel enumeration and discount optimization of
Section~\ref{sec:instrument-ladder} with its own predicted probabilities,
executes the first action and re-plans next month. Only GBM-lags and GBM-flat are
graded at both horizons. They form one solver class, differing only in their
history inputs, so their horizon contrasts compare each solver with itself.

Horizons in Table~\ref{tab:solvers} are in months. CE denotes certainty-equivalent
chaining of predicted monthly probabilities. Training rows, calibration and
decision rules are detailed in Appendix~\ref{app:solvers}.

\begin{table}[H]
\centering
\caption{Graded solvers.}
\label{tab:solvers}
\footnotesize
\setlength{\tabcolsep}{3pt}
\begin{tabular}{@{}>{\raggedright\arraybackslash}p{0.13\linewidth}>{\raggedright\arraybackslash}p{0.26\linewidth}>{\raggedright\arraybackslash}p{0.20\linewidth}>{\raggedright\arraybackslash}p{0.22\linewidth}>{\centering\arraybackslash}p{0.09\linewidth}@{}}
\toprule
Solver & Information & Model or policy & Decision rule & Horizon \\
\midrule
GBM-lags & Public state, current reading, six months of readings and channels & Two calibrated GBM classifiers & Myopic or sequential, reading and history features frozen during search & 1, 6 \\
GBM-flat & Public state and current reading & As GBM-lags & As GBM-lags, without history & 1, 6 \\
CE-Seq & Current state and reading, full pre-decision history including account states & Recurrent belief with two calibrated heads & Sequential, future readings masked & 6 \\
GBM-cash & Public state and current reading & GBM cash predictor, Tweedie loss & Myopic & 1 \\
LLM-agent & Delinquency ageing and balance & Fixed rules written from censored logs & Rules applied monthly & 1 \\
\bottomrule
\end{tabular}
\end{table}

The recurrent configurations in Section~\ref{sec:results-memory} are predictive
memory encoders, evaluated separately from these policy-value comparisons.

\section{Results}
\label{sec:results}

Closed-loop contrasts are reported with 95\% paired bootstrap intervals over loans, with
10{,}000 resamples that draw the same loans for every solver. These are percentile
intervals, with no correction for multiple comparisons. Each fitted solver uses one
training world and one fit, so the closed-loop intervals do not include variation from
retraining. Contrasts between graded solvers use the fixed cohort optimum as the
normalizer. Contrasts of the declared intervention sequence and of the two fitted
refinements of GBM-lags (Section~\ref{sec:results-base}) recompute it within each
resample (Appendices~\ref{app:board} and~\ref{app:ladder}). Monte Carlo references carry
per-reference standard errors (Section~\ref{sec:instrument-ladder}). A point is one
percentage point of the member's upper reference.

For the predictive measurements, probe scores (Table~\ref{tab:probes}) are out-of-fold
means over three cross-validation folds. The recurrent configurations of
Section~\ref{sec:results-memory} report the mean and population standard deviation over
five training seeds on Mem-4 and two on the other members. A point of $R^2$ is $0.01$.

\subsection{Base}
\label{sec:results-base}

\begin{figure}[!ht]
\centering
\includegraphics[width=\linewidth]{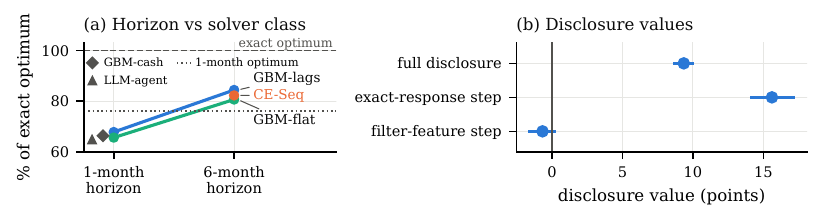}
\caption{Closed-loop performance on Base.}
\label{fig:board}
\end{figure}

On Base, within the GBM class, the only class graded at both horizons,
extending the planning horizon from one month to six raises the score of
GBM-lags by 16.62 points ($[15.31, 18.02]$) and that of GBM-flat by 15.06
($[13.54, 16.66]$) of the exact optimum.
At a fixed horizon, the largest
point estimate is 3.75 ($[2.75, 4.78]$), between GBM-lags and GBM-flat.
Across fitted solver classes, the largest contrast is 2.12 ($[0.82, 3.34]$),
by which GBM-lags exceeds CE-Seq despite the latter's access to earlier
account states (Fig.~\ref{fig:board}, Appendix~\ref{app:board}).

The connected
pairs in Fig.~\ref{fig:board}(a) are the within-class horizon contrasts,
unconnected markers are solvers graded at one horizon, and the dotted rule
marks the one-month optimum. LLM-agent, graded once at the one-month horizon,
reaches 64.74\% of the exact optimum, 0.85 points ($[0.05, 1.73]$) below the
closest fitted solver, GBM-flat at one month (Table~\ref{tab:board-contrasts}).

Full disclosure, whose endpoint is the filter planner of Table~\ref{tab:references},
improves the score of GBM-lags by 9.35 points ($[8.68, 10.02]$) of the 15.60 ($[14.36,
17.01]$) separating it from the exact optimum. The declared intervention sequence
(Fig.~\ref{fig:board}b, Appendix~\ref{app:ladder}) replaces the fatigue interpolation,
the state features, their propagation during search and the fitted response. The
response replacement uses each loan's true hidden traits. Replacing the history features
by the reference filter's features changes the score by $-0.68$ points, with an interval
that includes zero. The largest step, $+15.61$ points ($[14.14, 17.16]$), replaces the
fitted response after the preceding changes. It exceeds the full gain because the
exact-fatigue, filter-feature and rolled-belief steps together lower the score by 6.26
points. The full gain and two steps are shown with 95\% intervals, with step values
conditional on the sequence and fitted response.

A linear model fitted separately for each contact channel predicts the hidden score from
the current reading and the last six months of readings and actions, with a row-weighted
mean out-of-sample $R^2$ of $0.56$ ($0.06$ when a single model is fitted across the
three channels). This oracle-side diagnostic is trained on the hidden score, whereas the
solvers learn from repayment labels. The structured estimator it motivated scores
$-1.73$ points ($[-2.60, -0.86]$) against GBM-lags, while an 11-lag window scores
$+0.63$ ($[-0.12, +1.47]$). These two fitted refinements use the same logs and cohort as
GBM-lags, and neither establishes an improvement (Appendix~\ref{app:ladder}).

These comparisons leave the source of the remaining gap unresolved. Contact
exploration randomizes actions with $\varepsilon = 0.03$ only in eligible states
and is held fixed across the comparisons, so its contribution is not identified.

For a fixed plan, certainty-equivalent chaining inflates the filter-based
planner's self-reported value, the value it computes for that plan, by
approximately 11\% relative to the value of the same plan integrated jointly
under the filter's belief (95\% loan-bootstrap interval $[10.8, 11.6]\%$).
On Base, this estimate uses 4{,}000 loans and 512 draws per loan and is
conditional on one geometry and the specified filter belief, with posterior
draws held fixed when resampling loans. Graded and self-reported values are
reported separately throughout.

\subsection{Memory}
\label{sec:results-memory}

\begin{figure}[!ht]
\centering
\includegraphics[width=0.78\linewidth]{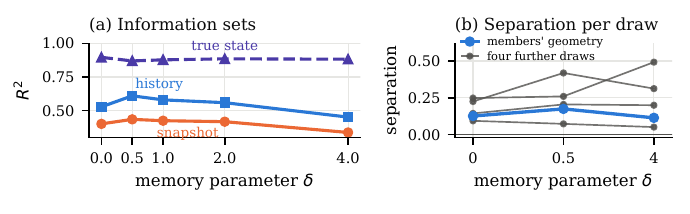}
\caption{Memory measurements.}
\label{fig:dial}
\end{figure}

The memory sweep compares the snapshot, history and true-state probe scores across
values of $\delta$ (Fig.~\ref{fig:dial}a). It uses pilot worlds, smaller unreleased
worlds of 100{,}000 borrowers over 12 months generated on the members' geometry. The
released members correspond to $\delta \in \{0, 0.5, 4\}$. On this geometry, the
history-minus-snapshot separation is $+0.127$ at $\delta = 0$, $+0.175$ at $\delta =
0.5$ and $+0.114$ at $\delta = 4$. At $\delta = 4$, headroom is $+0.430$ and the
snapshot score is $0.338$. These fitted contrasts differ from the margin to the
information reference of Eq.~\ref{eq:bound} reported below.

The geometry sweep repeats these measurements across five independent draws of
the world's matrices (Fig.~\ref{fig:dial}b). The first ordering clause requires separation at $\delta = 0.5$ to exceed
that at $\delta = 0$. The second requires greater headroom and no higher
snapshot score at $\delta = 4$ than at $\delta = 0$. Each clause holds in
four of five draws. The snapshot score falls from $\delta = 0$ to $4$ in five of five,
and the within-world separation is positive in all fifteen
draw--$\delta$ cells.
Worlds across $\delta$ are not matched populations. Each
history-minus-snapshot contrast uses the same rows within its own world.
Snapshot scores also vary across seeds and geometry draws. These runs do not
isolate an effect of $\delta$ on that score.

To test how learned memories use the history measured by the probes, each
recurrent configuration uses a gated recurrent unit (GRU) or long short-term
memory (LSTM) network to compress a loan's full history of readings and actions
into a vector. A small network combines this vector with the current month's
snapshot to predict the probes' target. Its out-of-fold $R^2$ minus the history
probe's score defines its memory value above that probe.

\begin{figure}[!ht]
\centering
\begin{minipage}[t]{0.57\linewidth}\centering
\includegraphics[width=\linewidth]{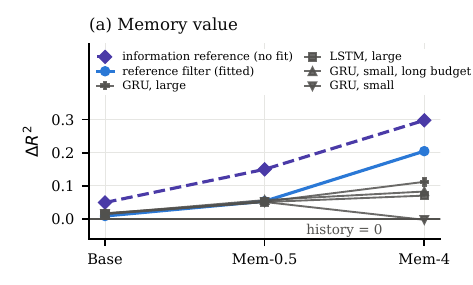}
\end{minipage}\hfill
\begin{minipage}[t]{0.40\linewidth}\centering
\includegraphics[width=\linewidth]{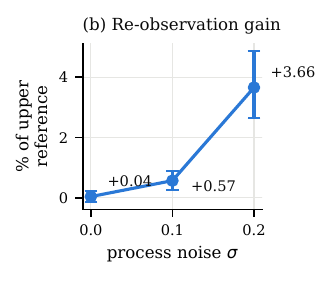}
\end{minipage}
\caption{Memory and re-observation.}
\label{fig:arch}\label{fig:reobs}
\end{figure}

On Mem-4, the information reference lies $0.298$ in $R^2$ above the history probe.
The recurrent configurations of Table~\ref{tab:arch} have gains of approximately
zero to 38\% of that margin at the stated training budgets, while the fitted
reference filter has a gain of 69\%
(Fig.~\ref{fig:arch}a, Appendix~\ref{app:archtable}). These percentages are a
normalization against the information reference, not fractions of attainable
predictive information.

On Mem-4, the long-budget configuration and the two large configurations reached the
epoch cap in every fold. Training was still improving in most folds of the two large
configurations and in about half the folds of the long-budget configuration. The
reported gains therefore describe the stated training budgets. Because capacity also
varies across configurations, the comparison does not isolate an architecture effect. At
$\delta \le 0.5$ the recurrent scores were close. The limited replication does not
establish equivalence between architectures.

\subsection{Process noise}
\label{sec:results-noise}

The re-observation gain is the value of re-planning every month on the updated filter
belief minus the value of executing the plan chosen at the decision point. On Base it is
approximately zero for the reference filter. It is positive on both noisy members, with
the larger gain at $\sigma = 0.2$~(Fig.~\ref{fig:reobs}b). The Base point uses a
closed-form comparison, while the noisy points use Monte Carlo evaluation. Measured
against each member's upper reference, re-planning with the filter belief gains 0.57
points at $\sigma = 0.1$ and 3.66 points at $\sigma = 0.2$. The corresponding gains in
expected recovered value per loan, in the world's currency units, are $+9.4$ ($[4.3,
14.8]$) and $+87.1$ ($[63.0, 116.1]$). These 95\% intervals resample loans together with
their 24 paths and are conditional on each member's evaluation world and geometry. The
two noisy members also differ in base rate, so this comparison does not isolate the
effect of $\sigma$ alone.

\subsection{Predictive comparison with real data}
\label{sec:results-transport}

The predictive measurements can also be extended to observed loan outcomes
by testing whether history improves prediction. The real-data comparison uses 20{,}677 delinquency spells, or episodes of
delinquency, constructed from Fannie~Mae's Single-Family Loan Performance
Data~\citep{fanniemae2024sf}, freely available upon registration. Here, cure means a return
to current status within six months. Adding the pre-entry block (twelve months
of delinquency states, prior-spell count, months since the previous cure) raises
the area under the receiver operating characteristic curve (AUC) for cure
prediction by $+0.041$ ($[0.031, 0.050]$). The interval resamples loans within
folds and is conditional on the fitted models and partition. It uses the same
mean-over-folds statistic as the point estimate, with 10{,}000 resamples.
Sensitivity to the partition is reported in Appendix~\ref{app:transport-table}.

The corresponding AUC gains in the synthetic worlds span $+0.010$ to $+0.013$, with all three
intervals excluding zero (Appendix~\ref{app:transport-table}). The real
portfolio's estimate lies above every synthetic interval. The two sides differ
in observations and targets. The real side is a mortgage portfolio observed
without contact actions. Its label is cure within six months, whereas the
synthetic label is recovery within nine months. Its history block also differs
from the synthetic readings and actions. The comparison concerns prediction. It
does not validate the effects of collection policies.

\section{Discussion}
\label{sec:discussion}

DECODE tests how specified changes to information and planning affect solver
performance under known dynamics. The results support interpretations tied to
the inputs, decision rules and grading objective used in each comparison.

On Base, extending lookahead within the same six-month evaluation window improves value
more than the tested solver differences at a fixed horizon, for the two GBMs tested at
both horizons. Costs, fatigue and state transitions can explain why current choices
affect later returns, although their individual contributions were not isolated. Solver
contrasts also change information inputs, and Base was selected for a gap between the
exact and one-month optima on a training-world prefix. The declared intervention
sequence recovers most of the remaining gap, with effects conditional on the fitted
response and preceding steps. Some steps lower value, and the filter-feature interval
includes zero. Since the filter omits pre-decision repayment outcomes, its small step
cannot establish that inference is generally unimportant. The final replacement supplies
the true response and hidden traits, combining information with model knowledge.
Separating an information penalty from a learning gap would require an optimum under the
solver's information constraints, which is not computed
here~\citep{ouerdane2026groundtruth}. Neither fitted refinement establishes an
improvement. The oracle-side readout predicts a hidden score, a different task from
learning repayment probabilities. Exploration, specification and extrapolation remain
possible explanations of the residual gap. The logging design is unchanged, so these
experiments do not isolate exploration as its cause.

History can help prediction even at zero memory parameter because readings remain
partial and noisy and action recency still affects the state. The parameter $\delta$
changes how effects depend on state and earlier actions. The history length supplied to
the solver, its representation and planning lookahead are separate design choices. A
peak in fitted history-minus-snapshot separation does not establish an optimum of memory
or data complexity. It may reflect a balance between historical information and the
predictor's ability to use it, a mechanism these experiments do not isolate. Separation
is positive across all measured geometry cells, but the ordering clauses are not
universal. Unmatched cross-parameter worlds also prevent attributing snapshot changes to
$\delta$ alone. The filter uses known dynamic and observation equations, while recurrent
encoders learn a representation and several were still improving at the training cap.
Varying capacities and budgets prevent a general architecture ranking, including for the
single attention configuration (Appendix~\ref{app:transformer}). The information
reference does not bound the probes, so shares of its margin are not fractions of
attainable information. History also improves prediction in the real mortgage portfolio,
with an estimate above the synthetic intervals. Populations, covariates and labels
differ, including six-month cure against nine-month recovery. Agreement in direction
neither calibrates synthetic effect sizes nor validates the effects of collection
policies. Predictive gains alone do not establish better decisions.

The observed self-report inflation comes from chaining monthly mean repayment
probabilities, which omits their dependence under state uncertainty. It is
measured against joint integration of the same plan under the same filter
belief and concerns that valuation rule, plan and belief. Re-planning on updated
filter beliefs improves value on both noisy members.
Gaussian innovations can make an initial plan less appropriate, but the
contractive dynamics keep this adaptation within a stable process. Base's
near-zero gain is empirical, not an identity at zero process noise, since readings
remain noisy. Larger benefits under stronger or abrupt shocks remain a hypothesis.
Adaptation depends on changes being observed, changing the preferred action
and being handled by the model. Regime changes and altered repayment mechanisms
were not tested, and different base rates between noisy members prevent
attributing their difference in gains to $\sigma$ alone.

These findings support diagnoses conditional on the evaluated configurations. Graded
solvers are evaluated on Base only. The memory members support predictive measurements
and the noisy members a comparison between reference policies. On Base, fitted policy
comparisons use one geometry, one training world and one fit per solver. Intervals
reflect the 4{,}000-loan cohort, without retraining variability. Monte Carlo errors
quantify uncertainty in reference estimates, not policy optimality. Under noise,
true-state re-planning normalizes scores without bounding attainable value. The policy
values concern the first-recovery grading objective. GBM-lags and GBM-flat plan with the
generator's charge-off hazard, cash formula and fatigue dynamics, so their horizon
contrast measures planning with a partly known model. No offline reinforcement learning
method that learns a value function over the planning horizon was graded. Broader
reliability claims require replication across training samples and worlds.
\section{Related work}
\label{sec:related}

Several benchmark suites summarize solver performance through policy returns. Offline
reinforcement learning~\citep{levine2020offline} suites such as D4RL~\citep{fu2020d4rl}
standardize datasets and report policy returns from logs that often include more
exploratory data than deployed policies produce~\citep{qin2022neorl}. Off-policy
evaluation (OPE) benchmarks, COBS~\citep{voloshin2021cobs} and DOPE~\citep{fu2021dope},
compare value estimators against returns from running the policy.

Studies in model-based reinforcement learning also compare methods by the return of the
policy trained against a learned model~\citep{janner2019mbpo,yu2020mopo}. The
value-equivalence line~\citep{grimm2020value} judges models by the value-based planning
they support. Learned recurrent state-space models are graded by downstream
return~\citep{hafner2021dreamerv2}, with no reference belief to compare against. DECODE
separately evaluates changes to model inputs and planning. Its memory experiments
examine the memory component of the world-model decomposition~\citep{ha2018world}. They
compare sequence encoders under a common readout design.

The relative performance of memory architectures depends on the task's memory
length~\citep{ni2023transformers}, and recurrent model-free baselines perform on
par with specialized methods when their architecture and hyperparameters are
chosen carefully~\citep{ni2022recurrent}. For DECODE, the comparisons are
conditional on the memory parameter and the information used by each reference.

Planning under the released dynamics is a partially observable Markov decision
problem~\citep{kaelbling1998pomdp}. Work on value overestimation identifies biases
from the maximization step~\citep{thrun1993issues,vanhasselt2016double}, selection
among noisy estimates~\citep{smith2006optimizer}, and selecting a policy on the
sample that estimated its model~\citep{mannor2007bias}. The inflation measured
here comes from chaining marginal repayment probabilities instead of integrating
their joint effect on survival and cash flows. It is measured against the same
plan integrated jointly under the filter's belief.

For systems fitted on historical data, interpretation also depends on how the
logs were collected. Identification rests on the randomization the logging
design provides~\citep{bottou2013counterfactual}. Sensitivity analyses under
unobserved confounding give bounds that can widen with the
horizon~\citep{namkoong2020confounding}. Reinforcement learning has been deployed
for collections~\citep{abe2010debt}, evaluated by importance weighting on logged
outcomes with no reference optimum.

JUBILEE~\citep{prado2026jubilee} evaluates OPE under hidden assignment dependence,
using an initial policy intervention followed by a common behavior policy.
DECODE examines controlled information and planning changes in sequential decision
systems, including the response to specified information disclosures. Its disclosure
contrasts differ from the decision-theoretic value of information~\citep{howard1966}.

Diagnostic suites use tractable environments to study specific capabilities.
bsuite~\citep{osband2020bsuite} includes experiments scored by regret normalized between
random and optimal policies. MDP Playground~\citep{rajan2023mdpplayground} varies
hardness dimensions over toy environments with known optima, uses the optimum for reward
normalization, and summarizes learning with area under the training curve.
POPGym~\citep{morad2023popgym} isolates memory models on partially observable tasks and
rescales each environment so that the worst and best observed trials map to 0 and 1.
Memory Gym~\citep{pleines2023memorygym} reports success rates and executed-command
counts. Memory Maze~\citep{pasukonis2023memorymaze} reports episode scores against a
privileged-information oracle and probes agents' representations against the true maze
layout. DECODE compares its predictive scores with the information reference,
conditional on the reading model, and policy values with a specified filter-based
planner.

Within this diagnostic literature, a study of deep Q-learning uses exact tabular
solutions to disentangle estimator-internal error sources, characterizing one
algorithm class on a fixed domain set~\citep{fu2019diagnosing}. Its oracle-stopping
headroom concerns stopping during learning, while the self-report inflation
measured here concerns valuation of a fixed plan.

A complementary approach generates environments with
certified optima along a difficulty ladder~\citep{ibrahim2026converse},
reporting normalized optimality gaps and discounted regret in a fully
observed setting. \citet{ouerdane2026groundtruth} extend this approach to
partially observed tasks with certified optima on a sufficient information state.
Their linear-quadratic-Gaussian (LQG) reference separates an analytic information
penalty from the learning gap under the stated model assumptions.

DECODE's contribution concerns controlled interventions in solvers whose
repayment responses are fitted on partially observed collection logs. These
experiments vary selected information and planning inputs under the same grading
objective. No optimum under the solvers' information constraints is computed,
so the contrasts measure the effects of specified interventions rather than
separating information loss from learning error.

\section{Conclusion}
\label{sec:conclusion}

DECODE supports diagnostic experiments on partially observed sequential decision systems
through controlled changes to information and planning inputs. On Base, a member
selected for a large gap between the exact and one-month optima, extending the horizon
within a fitted solver raises its score by 15 to 17 points of the exact optimum. At a
fixed horizon, contrasts between the tested solvers have point estimates below 4 points.
The declared intervention sequence recovers nine of the 15.6 points separating GBM-lags
from the exact optimum. The effects of these steps are conditional on their order, the
response model and the information supplied.

A separate fixed-plan comparison checks the planner's internal valuation.
On Base, chaining marginal repayment probabilities inflates the filter-based
planner's self-report relative to joint integration of the same plan under the
same belief.

The diagnoses drawn from these experiments concern the specified input
substitutions, information sets and training budgets in one domain. They do not assign a unique cause to the total performance
gap. The generator and reference policies allow these conditions to be
varied and their consequences measured.

\section*{Acknowledgments}

The authors review all AI-assisted contributions and take responsibility
for the manuscript, code and reported results.

\subsection*{LLM statement}

Claude Fable 5 (Anthropic) assisted with software implementation, 
literature retrieval, analysis checks, manuscript drafting and revision.

Loan records, rewards and policy-value estimates are produced by the simulator
and evaluation code using the specified configurations and random seeds.
Reported statistics are computed by code from saved inputs and outputs.

\bibliographystyle{decode-tmlr}
\bibliography{paper}

\clearpage
\appendix

\section{Glossary}
\label{app:glossary}

The terms defined in the body of the paper are collected here for
reference, in alphabetical order (Table~\ref{tab:glossary}).

\begingroup
\footnotesize
\setlength{\tabcolsep}{4pt}
\begin{longtable}{@{}>{\raggedright\arraybackslash}p{0.24\linewidth}>{\raggedright\arraybackslash}p{0.72\linewidth}@{}}
\caption{Glossary.}\label{tab:glossary}\\
\toprule
Term & Meaning \\
\midrule
\endfirsthead
\toprule
Term & Meaning \\
\midrule
\endhead
\bottomrule
\endlastfoot
Base, Mem-0.5, Mem-4, Noise-0.1, Noise-0.2 & the five released environments, with parameter settings in Table~\ref{tab:members} \\
belief & the probability distribution of the hidden state given everything a solver has observed \\
certainty-equivalent (CE) & chaining monthly repayment probabilities without their cross-month dependence \\
closed-loop grading & grading in which each solver re-decides every month on the readings its own actions generate \\
decision point & month six of an evaluation loan, when a solver takes over after six months of public history, with six months left to decide \\
disclosure & the replacement of one input of a graded solver by a specified reference quantity \\
exact optimum & the value of the best plan computed with the true hidden state and borrower traits at the decision point, available on members without process noise \\
exploration rows & logged loan-months on which the logging policy drew a contact channel and a discount at random \\
filter belief & the reference filter's Gaussian approximation, ignoring repayment outcomes \\
gap & the difference between the exact optimum and a solver's value, in points \\
geometry & one random draw of the fixed matrices and vectors in a world's equations. The five released members share one geometry \\
headroom & the true-state probe score minus the history probe score \\
hidden score & the channel-specific hidden-state term $\kappa\, u_c^{\top} z_t$, with terms defined after Eq.~\ref{eq:logit} \\
hidden traits & three unobserved borrower factors and a regional economic-stress term entering repayment \\
information clauses & floors and ceilings, fixed before generation, on probe scores, and on Base also on values, that a member with an exact reference must pass \\
information reference & the analytic $R^2$ reference conditional on the specified reading model and its information set \\
logging policy & the fixed rule that generated the training logs \\
member & one released setting of the memory parameter, the process-noise scale and the monthly base rate \\
memory parameter $\delta$ & the scale of the rotation each contact applies to the hidden state, which makes contacts interact with the state \\
myopic solver & a solver that chooses each month the action with the highest expected value for that month alone \\
one-month optimum & the best policy that plans one month at a time with the true hidden state \\
oracle & the grader's side of the world, which holds the hidden state and the equations \\
pilot world & a smaller unreleased world, of 100{,}000 borrowers over 12 months, used for sweeps and construction checks \\
point & one percentage point of a member's upper reference \\
probe & the out-of-fold $R^2$ of a gradient-boosted model predicting the true repayment logit from one information set, snapshot, history, or true state (Table~\ref{tab:probes}) \\
process-noise scale $\sigma$ & the standard deviation of the monthly innovation in the hidden state \\
reading & the noisy two-dimensional linear observation of the hidden state \\
re-anchoring & regenerating a member with an adjusted base rate after its aggregate recovery fell outside the recovery-rate band \\
recovery-rate bands & ranges into which the share of loans recovered within one month and within nine months must fall \\
reference & a value computed from the world's equations, against which a solver's result is read \\
reference filter & the Gaussian reading-model recursion with fixed actions, ignoring repayment outcomes \\
reference ladder & the analytic and Monte Carlo references available under each member's dynamics \\
reference solver & a labeled copy of a graded solver that receives a disclosure and never competes \\
re-observation gain & the value of re-planning every month on the updated belief minus the value of executing the plan chosen at the decision point \\
rotation & the orthogonal matrix $Q_w = \exp(w\,\delta\,S)$ that turns the hidden state, with terms defined after Eq.~\ref{eq:transition} \\
self-reported value & the value a planner computes for its own plan \\
separation & the history probe score minus the snapshot probe score \\
sequential solver & a solver that plans the whole remaining horizon every month, executes the first action, and re-plans \\
solver & a decision system graded by the instrument, built from the training logs \\
solver class & a solver's architecture and estimator, which can be graded at more than one horizon \\
upper reference & the normalizer of each member, the exact optimum without process noise and the value of monthly re-planning with the true state under noise \\
\end{longtable}
\endgroup

\section{Members}
\label{app:members}

The five members of Section~\ref{sec:instrument} are listed in Table~\ref{tab:members}
with their transition model, their parameters, their monthly base rate, and the outcome
of their construction checks. All reports record passing recovery-rate bands, and the
three members with $\sigma=0$ also pass their information clauses. These clauses set
floors on the separation or the headroom of the probes of Table~\ref{tab:probes}, so
that the history or the hidden state carries information the current month lacks, and,
on the memory members, ceilings on what the current month alone predicts, so that the
reading does not leak the hidden state. On Base the clauses are also stated on values,
requiring that planning over the horizon matter, that the best plan vary across loans,
and that a solver without history stay well below the exact optimum. The first requires
the exact optimum to exceed the one-month optimum by 15 to 40 points, and Base measured
27.1 on a 100{,}000-borrower prefix of its training world. The history-free clause caps
a solver without history, planning over six months, at 85 points. This solver was fit
and graded on the same training-world prefix, where Base measured 59.2 before any solver
of Section~\ref{sec:results} was graded. The clauses on plan variation require that no
single channel sequence be optimal for more than 60\% of loans, where Base measured
25\%, and that the best of the twelve most frequent optimal channel sequences, applied
to every loan with a 20\% discount on contact months, stay at least 2 points below the
exact optimum, where Base measured 19.9 points. A coverage clause requires at least 50
logged rows in each of 27 cells formed by income terciles, three contact-fatigue levels
and three channel classes, and the smallest Base cell held 7{,}629 rows. The two memory
members also pass a per-loan consistency check, under which the plan optimum is never
below the one-month optimum, and on Base the same holds on every graded loan. The
thresholds of every clause are recorded with the release.

The release carries each member's machine-readable check report and a manifest recording
the configuration hash, the seeds, and the pinned generation batch size that
regeneration requires. Check criteria were fixed before generation, with one exception.
An information clause on Mem-0.5 failed and was replaced once before release by a clause
targeting the same mechanism, and the amendment is in the release records. Three members
were re-anchored, that is, regenerated with an adjusted base rate after their aggregate
recovery fell outside the recovery-rate band, two on pilot or check worlds before definitive
generation and one after its definitive world failed the check. The release contains
each member's final configuration and check report and, for Mem-4, the configuration and
check report of the definitive world that failed.

Two further verifications were run at
construction time on pilot worlds. With $\delta = 0$ the generator
reproduces bit for bit the output of its version without the memory
parameter, so the parameter changes nothing when it is off, and the
grader's replay of each client's hidden-state path under the logged
actions agrees with the generator to within $10^{-12}$.

\begin{table}[!ht]
\centering
\caption{Members and release checks.}
\label{tab:members}
\footnotesize
\begin{tabular}{llllll}
\toprule
Member & Transition model & Memory $\delta$ & Noise $\sigma$ & Base rate & Checks \\
\midrule
Base      & linear, deterministic     & 0   & 0   & 0.038 & pass, re-anchored \\
Mem-0.5   & action-rotation           & 0.5 & 0   & 0.038 & pass, one clause amended \\
Mem-4     & action-rotation           & 4   & 0   & 0.026 & pass, re-anchored \\
Noise-0.1 & linear, stochastic        & 0   & 0.1 & 0.038 & pass \\
Noise-0.2 & linear, stochastic        & 0   & 0.2 & 0.032 & pass, re-anchored \\

\bottomrule
\end{tabular}
\end{table}

\section{Solver training details}
\label{app:solvers}

The information inputs and decision rules of the graded solvers are summarized
in Table~\ref{tab:solvers}. GBM-lags and GBM-flat each fit two calibrated
classifiers using training-world months six onward. The contact classifier uses
exploration rows, while the no-contact classifier uses logged no-contact rows.
GBM-lags adds a six-month window of readings and channel codes. Both solvers keep
their reading features, and GBM-lags its history features, fixed at the decision
month during sequential search.

CE-Seq fits a recurrent belief model on the full training panel, summarizing the
account's observed history from its first month. Its contact head uses
exploration months and its no-contact head uses logged no-contact months, each
with its own isotonic calibrator. Here certainty-equivalent means chaining
predicted monthly repayment probabilities as fixed quantities along a plan.
This differs from planning with the state fixed at its conditional
mean~\citep{barshalom1974dual}. Future readings are masked during search. Its
pre-decision account-state sequence is read directly from the evaluation
database, in addition to the history supplied by the harness.

GBM-cash fits expected monthly cash with a Tweedie loss on all training-log rows,
using a temporal validation split. It chooses one-month actions. LLM-agent uses
state-conditional rules written once after a language model examined a censored
copy of the training logs, without access to the evaluation world. The rules
were frozen before evaluation and applied monthly.

\section{Declared intervention sequence}
\label{app:ladder}

The full disclosure of Section~\ref{sec:results-base} is reached from GBM-lags in four specified steps, each of which
changes one input or its numerical evaluation and is graded on the same 4{,}000 loans with one
shared-index bootstrap (Table~\ref{tab:ladder}). Values are percentages of the
exact optimum, and steps measure changes relative to the preceding row. Their
sum equals the full-disclosure gain by identity. The exact-fatigue step replaces
interpolation with exact evaluation of
the observed fatigue along each candidate path. It is a numerical control,
not a disclosure of hidden information. The filter-feature step has an interval
that includes zero, and propagating those features lowers value. The exact-response
step raises value after these preceding changes, exceeding the total gain.
These effects are conditional on the preceding steps and on the fitted
response. The remaining 6.25 points between full disclosure and the
exact optimum combine differences in information with the planner's
approximations. The filter ignores pre-decision repayment outcomes and
its planner chains marginal probabilities. The remaining distance therefore
does not measure state-inference error alone. The intervals of the sequence contrasts divide each
bootstrap resample's mean difference by that resample's mean optimum.

\begin{table}[!ht]
\centering
\caption{Declared intervention sequence on Base.}
\label{tab:ladder}
\footnotesize
\setlength{\tabcolsep}{4pt}
\begin{tabular}{@{}l>{\raggedright\arraybackslash}p{0.32\linewidth}rrr@{}}
\toprule
Solver & Change from the row above & Value & Step & 95\% interval \\
\midrule
GBM-lags & none & 84.40 & & \\
exact fatigue & exact per-path fatigue instead of interpolation & 83.65 & $-0.75$ & $[-1.18, -0.34]$ \\
filter features & reference-filter features instead of history & 82.97 & $-0.68$ & $[-1.60, +0.22]$ \\
rolled belief & filter belief propagated instead of frozen & 78.14 & $-4.83$ & $[-6.29, -3.43]$ \\
exact response & true response with hidden traits instead of fitted response & 93.75 & $+15.61$ & $[14.14, 17.16]$ \\
\midrule
full disclosure & all four changes together & & $+9.35$ & $[8.68, 10.02]$ \\
\bottomrule
\end{tabular}
\end{table}

Measured directly against GBM-lags, the three intermediate reference solvers of the
declared intervention sequence score $-0.75$ ($[-1.18, -0.34]$), $-1.43$ ($[-2.37,
-0.55]$) and $-6.26$ ($[-7.77, -4.80]$). They serve as controls in that sequence, rather
than additional fitted competitors. Exact fatigue changes the numerical evaluation of a
public state, whereas the filter replacements change the state information.

The two fitted refinements of Section~\ref{sec:results-base} are a structured
estimator and an 11-lag window. The structured estimator replaces the history
features with a linear correction in logit space, fitted on top of a history-free
GBM. Its specification drew on the oracle-side diagnostic, and its search uses
exact per-path fatigue. It scores $-1.73$ ($[-2.60, -0.86]$) against GBM-lags.
Against the exact-fatigue control, its difference is $-0.98$ points. The 11-lag
window adds public history and has an interval that includes zero at $+0.63$
($[-0.12, +1.47]$). These comparisons do not establish an improvement over
GBM-lags.

A separate experiment compares response-component substitutions under its own protocol,
with the fixed cohort optimum as normalizer and spans not additive to the declared
intervention sequence. In that experiment, propagating the exact contact component
improves value, whereas rolling filter features into the fitted response lowers it in
the declared intervention sequence. The fitted inputs also differ, so this contrast does
not isolate order alone. Relative to the exact-fatigue control, substituting the exact
contact response with frozen features changes value by $+6.7$ points ($[+5.7, +7.8]$),
propagating its effect across months adds $+3.3$ ($[+2.3, +4.0]$), and substituting the
exact response on no-contact months, the organic component, changes value by $-0.32$
($[-0.63, -0.05]$). The no-contact model is fit on observational rows. These contrasts
do not isolate the effect of randomizing no contact.

\section{Memory values}
\label{app:archtable}

The memory scores of Fig.~\ref{fig:arch}a on Mem-4 are tabulated in
Table~\ref{tab:arch} as gains above the history probe at the stated training
budgets. The recurrent LSTM~\citep{hochreiter1997lstm} and
GRU~\citep{cho2014gru} variants report means with population standard deviations
over five training seeds. The share column divides each gain by the $0.298$
margin between the history probe and the information reference, rather than by
attainable information. The final column records whether the epoch cap was reached.

On Base's training world, the history probe reaches $R^2$ of $0.624$, a
fitted pipeline on the reference filter's outputs $0.633$, and a fitted
recurrent memory $0.639$. These scores lie below the $0.674$ information
reference, which does not bound them
(Section~\ref{sec:instrument-ladder}). These levels are from the training
world and are not comparable to the pilot-world sweep of
Fig.~\ref{fig:dial}.

\begin{table}[!ht]
\centering
\caption{Memory value on Mem-4.}
\label{tab:arch}
\footnotesize
\begin{tabular}{lrrl}
\toprule
Configuration & $\Delta R^2$ vs history & share & at epoch cap \\
\midrule
recurrent (GRU), small & $-0.002 \pm 0.003$ & $\approx 0$ & no stopping log \\
recurrent (GRU), small, long budget & $+0.083 \pm 0.002$ & 27.9\% & yes \\
recurrent (LSTM), large & $+0.071 \pm 0.006$ & 23.7\% & yes \\
recurrent (GRU), large & $+0.112 \pm 0.002$ & 37.7\% & yes \\
reference filter (fitted) & $+0.205$ & 68.8\% &  \\
information reference (no fit) & $+0.298$ & 100\% &  \\

\bottomrule
\end{tabular}
\end{table}

\section{Attention}
\label{app:transformer}

Alongside the recurrent configurations of Table~\ref{tab:arch}, one
attention-based sequence model was evaluated with the same readout head,
folds, and data. The encoder is a two-layer transformer~\citep{vaswani2017attention} of
model width 64 with four heads, causal masking, and learned positions. Its epoch cap
and patience match the standard recurrent configurations' at a lower learning
rate. In Table~\ref{tab:tfm}, the upper block compares training budgets and the
lower block reports memory value above the history probe for each member. One
recurrent configuration of Table~\ref{tab:arch} ran a longer
budget, so shares credited to it and the values reported here are
not comparable at equal optimization effort. Each member cell
trains three folds per seed. Dispersion is the population standard deviation
over seed means. The two-seed cells include individual values, showing opposite
signs on Mem-0.5. Every transformer fold reached the epoch
cap, twelve of the fifteen Mem-4 folds were still improving there, and
three had peaked before the cap. These measurements are conditional
on the stated training budget.
A single small configuration at one budget does not establish a general
ranking of attention-based and recurrent memories. The per-fold stopping logs
are recorded with the measurement.

\begin{table}[!ht]
\centering
\caption{Attention model budgets and memory values.}
\label{tab:tfm}
\begin{tabular}{lccc}
\toprule
configuration & epoch cap & patience & learning rate \\
\midrule
attention (this appendix) & 30 & 5 & $3\times10^{-4}$ \\
recurrent (Table~\ref{tab:arch}) & 30 & 5 & $10^{-3}$ \\
recurrent, long budget (Table~\ref{tab:arch}) & 100 & 15 & $10^{-3}$ \\
\midrule
member & seeds & memory value & per-seed \\
\midrule
Base & 2 & $-0.021$ & $-0.022$, $-0.020$ \\
Mem-0.5 & 2 & $0.000$ & $-0.008$, $+0.008$ \\
Mem-4 & 5 & $-0.090 \pm 0.005$ & \\
\bottomrule
\end{tabular}
\end{table}

\section{Closed-loop values}
\label{app:board}

Every closed-loop value on Base, including
the exact and one-month optima, a random policy, and the single-observation LLM-agent
reported in Section~\ref{sec:results-base}, is listed in
Table~\ref{tab:board}. Values are levels expressed as percentages of the exact
optimum. Every solver row is also plotted in Fig.~\ref{fig:board}(a), and the
random policy provides the comparison floor. Comparisons between rows use the
paired contrasts of Table~\ref{tab:board-contrasts} under the bootstrap protocol
of Section~\ref{sec:results}.

\begin{table}[!ht]
\centering
\caption{Closed-loop values on Base.}
\label{tab:board}
\footnotesize
\begin{tabular}{lrr}
\toprule
Solver & 1-month horizon & 6-month horizon \\
\midrule
GBM-lags & 67.78 & 84.40 \\
GBM-flat & 65.59 & 80.65 \\
CE-Seq & & 82.28 \\
GBM-cash & 66.32 & \\
LLM-agent & 64.74 & \\
random policy & \multicolumn{2}{c}{29.11} \\
1-month optimum & 76.13 & \\
exact optimum & & 100 \\
\bottomrule
\end{tabular}
\end{table}

The paired contrasts in Table~\ref{tab:board-contrasts} are in points of the
exact optimum, with 95\% paired bootstrap intervals over the 4{,}000 evaluation
loans. The 10{,}000 resamples draw the same loans for every solver and use the
fixed cohort optimum as the normalizer.

\begin{table}[!ht]
\centering
\caption{Paired contrasts on Base.}
\label{tab:board-contrasts}
\footnotesize
\begin{tabular}{lrr}
\toprule
Contrast & Points & 95\% interval \\
\midrule
GBM-lags, six months minus one month & $+16.62$ & $[15.31, 18.02]$ \\
GBM-flat, six months minus one month & $+15.06$ & $[13.54, 16.66]$ \\
GBM-lags minus GBM-flat, six months & $+3.75$ & $[2.75, 4.78]$ \\
GBM-lags minus CE-Seq, six months & $+2.12$ & $[0.82, 3.34]$ \\
CE-Seq minus GBM-flat, six months & $+1.63$ & $[0.17, 3.21]$ \\
GBM-lags minus GBM-flat, one month & $+2.19$ & $[1.32, 3.15]$ \\
GBM-lags minus GBM-cash, one month & $+1.46$ & $[0.61, 2.44]$ \\
GBM-flat minus LLM-agent, one month & $+0.85$ & $[0.05, 1.73]$ \\
exact optimum minus GBM-lags, six months & $+15.60$ & $[14.36, 17.01]$ \\
exact optimum minus one-month optimum & $+23.87$ & $[22.43, 25.40]$ \\
\bottomrule
\end{tabular}
\end{table}

\section{Predictive gains from pre-entry history}
\label{app:transport-table}

The AUC gain of Section~\ref{sec:results-transport} is reported in
Table~\ref{tab:transport} as the out-of-fold $\Delta$AUC from adding pre-entry history.
The real portfolio and the released members use the same estimator and a five-fold,
loan-grouped evaluation policy. The real-side outcome is cure within six months, while
the synthetic label is recovery within nine. The intervals use 10{,}000 resamples of
loans within folds and condition on the fitted models and partition. The final column
gives the minimum and maximum over 20 random loan-grouped partitions with refitted
models. This descriptive range is not a confidence interval.

The real-data source comprises quarterly files from 2000Q1 through 2025Q4. The data
license restricts redistribution, so the release records a cohort fingerprint rather
than loan identifiers. The identifier list is withheld under the data license. The
real-data calculation was reproduced with scikit-learn 1.7.2. Reproduction under the
release's pinned 1.9.0 environment has not been verified.

\begin{table}[!ht]
\centering
\caption{Predictive gains from pre-entry history.}
\label{tab:transport}
\footnotesize
\begin{tabular}{lrrr}
\toprule
Data & $\Delta$AUC & 95\% interval & Partition range \\
\midrule
Fannie Mae spells (real, $n = 20{,}677$) & $+0.041$ & $[0.031, 0.050]$ & $[0.030, 0.045]$ \\
Base      & $+0.013$ & $[0.010, 0.016]$ & $[0.011, 0.014]$ \\
Mem-0.5   & $+0.013$ & $[0.010, 0.016]$ & $[0.012, 0.017]$ \\
Mem-4     & $+0.010$ & $[0.007, 0.013]$ & $[0.008, 0.013]$ \\
\bottomrule
\end{tabular}
\end{table}

\end{document}